\documentclass[11pt,a4paper]{article}

\usepackage[margin=1in]{geometry}
\usepackage{amsmath,amssymb,mathtools}
\usepackage{graphicx}
\usepackage{booktabs,tabularx,array,multirow}
\usepackage{enumitem}
\usepackage{caption}
\usepackage{xcolor}
\usepackage[hidelinks]{hyperref}
\usepackage{microtype}

\graphicspath{{figures/}}
\setlist{nosep,leftmargin=*}
\newcolumntype{Y}{>{\centering\arraybackslash}X}
\newcommand{\ours}{YOLO12-MambaScan}

\title{\textbf{YOLO12-MambaScan: An Efficient Object Detector with High-Frequency Enhancement and State-Space Modeling}}
\author{Hao Wang\\BDNRC\\\texttt{wh1090220084@163.com}}
\date{}

\begin{document}
\maketitle

\begin{abstract}
The rapid development of unmanned aerial vehicle (UAV) technology has made aerial-image object detection increasingly important for natural-resource monitoring, traffic management, and disaster response. Detecting small objects in aerial images remains difficult because objects occupy very few pixels, high-frequency cues are easily lost, and global context is hard to model in cluttered scenes. Existing detectors often retain insufficient edge, corner, and texture information. We propose \ours, an aerial-image detector built on the YOLO12 architecture. The model combines a triple-path high-frequency enhancement convolution module (TriPathHFConv), receptive-field coordinate-attention convolution (RFCAConv), and a Mamba-based global-context module. On VisDrone, at an input resolution of $960\times960$, \ours achieves 60.0\% mAP@50 and 38.6\% mAP@50:95, demonstrating a favorable accuracy--efficiency trade-off for small-object detection. The benchmark and dataset protocol follow the VisDrone challenge setup~\cite{zhu2018visdrone}.
\end{abstract}

\textbf{Keywords:} object detection; UAV aerial imagery; YOLO12; high-frequency feature enhancement; state-space model; small-object detection

\section{Introduction}
UAV aerial-image object detection is an important computer-vision problem with applications in urban management, agricultural monitoring, and border patrol. Unlike conventional photographs, aerial images are usually captured from a top-down viewpoint, contain large scale variation dominated by small objects, include many visually similar background distractors, and are affected by strong changes in illumination and weather. These properties make generic object detectors less reliable in aerial scenes.

Studies of small-object recognition show that discriminative evidence is concentrated in high-frequency components such as edges, corners, and fine textures. Standard convolution tends to smooth these components during feature extraction. HRMamba-YOLO combines a high-resolution feature pyramid with Mamba blocks and reports improved UAV small-object detection, highlighting the value of preserving high-frequency information while modeling global context~\cite{liu2025hrmamba}.

The YOLO family is widely used because it offers a strong speed--accuracy balance. YOLO12 adopts an attention-centric design with regional attention and a residual efficient layer aggregation network (R-ELAN), while FlashAttention reduces memory traffic~\cite{tian2025yolov12,dao2022flashattention}. Nevertheless, its generic design leaves room for improvement on aerial images: small objects require finer high-frequency preservation, shallow details are underused, and the background and scale variation of UAV scenes are not explicitly addressed.

To address these issues, we introduce \ours, an improved YOLO12 detector designed for aerial imagery. Our contributions are:
\begin{enumerate}
  \item RFCAConv, which shifts attention from conventional spatial features to receptive-field features, alleviating the limitations of shared convolution kernels during downsampling.
  \item TriPathHFConv, which explicitly preserves high-frequency details through basic depthwise, dilated depthwise, and high-frequency enhancement paths.
  \item MambaBlock, which uses selective state-space modeling to capture global context with linear complexity~\cite{gu2023mamba}.
  \item GSPN\_SCSA, which jointly models spatial and channel attention and improves information flow between the neck and detection head.
  \item STEL, a small-target enhancement layer with a new P2/4 prediction scale and a dual-branch feature path.
\end{enumerate}

On VisDrone, the complete model reaches 60.0\% mAP@50 and 38.6\% mAP@50:95 while retaining a practical computational cost.

\section{Related Work}
\subsection{UAV Aerial-Image Object Detection}
VisDrone is one of the most influential benchmarks for UAV detection. It contains images collected in 14 cities in China, covers urban and rural environments, and provides more than 2.6 million bounding-box annotations for ten categories, including pedestrians, cars, and bicycles~\cite{zhu2018visdrone}. Early work generally fine-tuned generic detectors such as Faster R-CNN~\cite{ren2015faster} and YOLO~\cite{redmon2016yolo}. Their performance is limited when small objects dominate and the background is cluttered. Recent aerial-specific methods include FBRT-YOLO, which uses a feature-complementary mapping module and a multi-kernel perception unit, and HRMamba-YOLO, which combines high-resolution features with Mamba~\cite{wang2025fbrt,liu2025hrmamba}.

\subsection{High-Frequency Feature Enhancement}
High-frequency cues are critical for small-object detection. Frequency-assisted Mamba linear attention modules fuse frequency- and spatial-domain features for multi-scale scenes~\cite{wang2025mfm}. SET (Spectral Enhancement for Tiny Object Detection) reports that tiny objects can become less salient after feature encoding and studies how selective removal or enhancement of high-frequency information affects recognition~\cite{li2024set}. Multi-Scale Dilated Attention (MSDA) combines dilated convolution and attention to capture local context at multiple scales. Together, these studies motivate the explicit high-frequency path in TriPathHFConv.

\subsection{Evolution of YOLO Detectors}
YOLO detectors formulate detection as a single-stage prediction problem~\cite{redmon2016yolo}. Later versions improved the backbone, label assignment, and training recipe. YOLOv9 introduced programmable gradient information~\cite{wang2024yolov9}; YOLOv10 removed non-maximum suppression from the end-to-end pipeline~\cite{wang2024yolov10}; and YOLO12 emphasizes attention, regional processing, R-ELAN, and efficient attention kernels~\cite{tian2025yolov12}. Generic YOLO models still face a tension between high-resolution spatial detail and global context on UAV images. \ours addresses this tension with dedicated high-frequency modules and Mamba-based context modeling.

\section{Method}
\subsection{Overall Architecture}
\begin{figure}[htbp]
  \centering
  \includegraphics[width=0.82\linewidth]{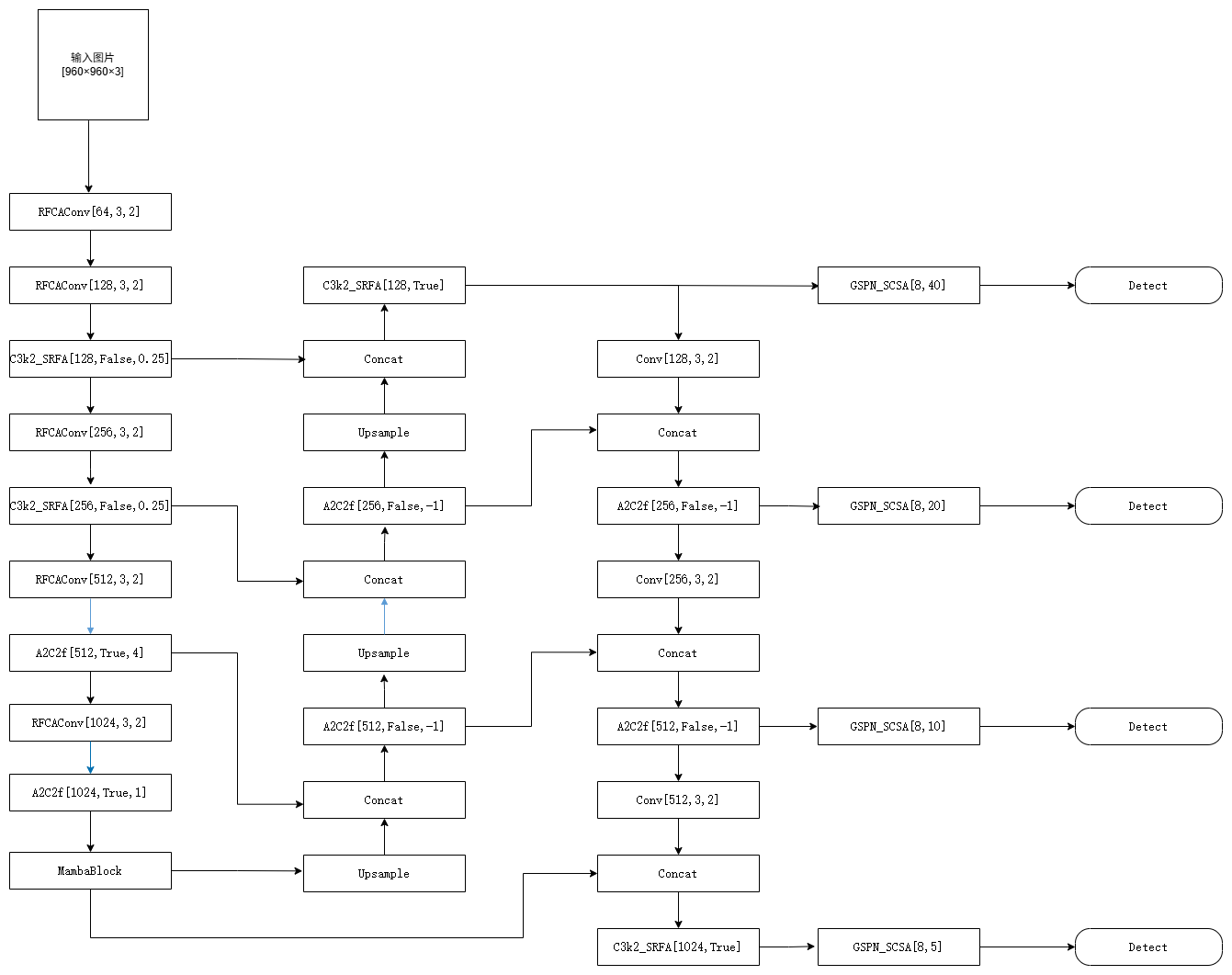}
  \caption{Overall architecture of YOLO12-MambaScan.}
  \label{fig:architecture}
\end{figure}

The overall architecture of \ours is shown in Fig.~\ref{fig:architecture}. It contains a backbone, a neck, and a detection head. In the shallow backbone (layers 2--4), TriPathHFConv replaces the original convolution layers. Standard downsampling layers are replaced with RFCAConv, and MambaBlock is inserted in the deepest stage. GSPN\_SCSA is placed between the neck and the head to improve multi-scale perception. The YOLO12 decoupled head is retained, while STEL adds a P2/4 high-resolution prediction branch for very small objects.

\subsection{Receptive-Field Coordinate-Attention Convolution (RFCAConv)}
In aerial detection, the shared-kernel assumption of standard convolution can suppress location-specific evidence from small objects. RFCAConv combines receptive-field attention (RFA) with coordinate attention (CA). Grouped convolution first generates receptive-field features whose support adapts to the kernel size. For an input feature map $X$, the core computation is
\begin{align}
F_{\mathrm{rf}} &= \operatorname{ReLU}\!\left(\operatorname{Norm}\left(g^{k\times k}(X)\right)\right),\\
A_{\mathrm{rf}} &= \operatorname{Softmax}\!\left(g^{1\times1}\left(\operatorname{AvgPool}(F_{\mathrm{rf}})\right)\right),\\
F &= A_{\mathrm{rf}}\odot F_{\mathrm{rf}},
\end{align}
where $g^{k\times k}$ is a grouped convolution, $\operatorname{AvgPool}$ is global average pooling, and $A_{\mathrm{rf}}$ is a receptive-field attention map. The coordinate formulation preserves horizontal and vertical positional information. Compared with standard convolution, RFCAConv adds approximately 1.5\% parameters and 2.3\% FLOPs while improving feature retention during downsampling. We use it at the P1/2, P2/4, P3/8, P4/16, and P5/32 scale transitions. RFCAConv complements TriPathHFConv: the former strengthens receptive-field representation during downsampling, whereas the latter preserves high-frequency detail during feature extraction.

\begin{figure}[htbp]
  \centering
  \includegraphics[width=0.72\linewidth]{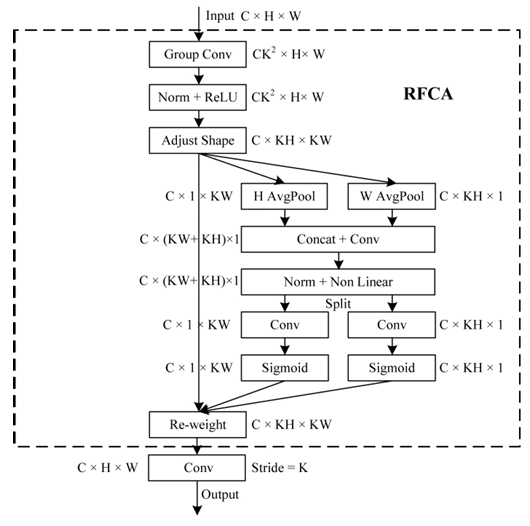}
  \caption{RFCAConv structure.}
  \label{fig:rfca}
\end{figure}

\subsection{Triple-Path High-Frequency Enhancement Convolution (TriPathHFConv)}
The edges, corners, and textures of distant pedestrians and vehicles are mainly high-frequency components. TriPathHFConv uses three parallel paths to preserve these details while retaining a broad receptive field.

The basic depthwise path captures local patterns:
\begin{equation}
X_1=\sigma\!\left(\operatorname{BN}_1\left(\operatorname{DWConv}(X)\right)\right),
\end{equation}
where $\sigma$ is SiLU. The dilated depthwise path enlarges the receptive field without a proportional parameter increase:
\begin{equation}
X_2=\sigma\!\left(\operatorname{BN}_2\left(\operatorname{DWDConv}(X_1)\right)\right).
\end{equation}
The high-frequency path predicts a content-adaptive attention map:
\begin{align}
A_{\mathrm{hf}} &= \operatorname{Sigmoid}\!\left(\operatorname{Conv}_{1\times1}\left(\operatorname{Conv}_{3\times3}(X)\right)\right),\\
X_3 &= \sigma\!\left(\operatorname{BN}_3\left(X\odot A_{\mathrm{hf}}\right)\right).
\end{align}
The path outputs are concatenated and projected:
\begin{align}
X_{\mathrm{concat}} &= \operatorname{Concat}(X_1,X_2,X_3),\\
Y &= \sigma\!\left(\operatorname{BN}_4\left(\operatorname{Conv}_{1\times1}(X_{\mathrm{concat}})\right)\right).
\end{align}
This design provides feature complementarity, information completeness, and adaptive channel reweighting. It simultaneously captures local detail, multi-scale context, and high-frequency evidence, avoiding the fixed-filter limitation of hand-designed enhancement operators.

\begin{figure}[htbp]
  \centering
  \includegraphics[width=0.48\linewidth]{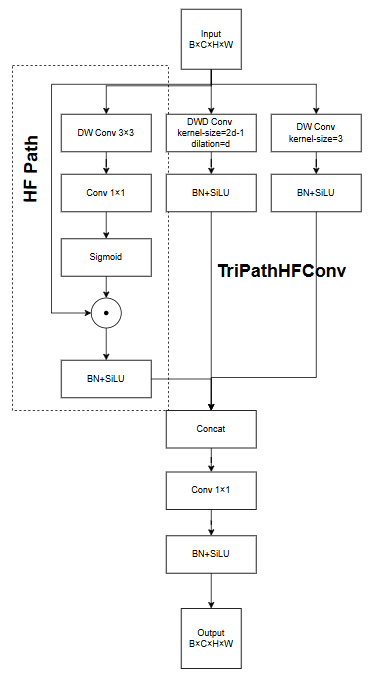}
  \caption{TriPathHFConv structure.}
  \label{fig:tripath}
\end{figure}

\subsection{Mamba-Based Global-Context Enhancement (MambaBlock)}
Self-attention captures long-range dependencies but has quadratic complexity with respect to the number of tokens. Mamba uses selective state-space modeling to provide global interaction with linear complexity~\cite{gu2023mamba}. We therefore insert MambaBlock at the deepest backbone stage, where features have high semantic abstraction and lower spatial resolution.

For $X\in\mathbb{R}^{B\times C\times H\times W}$, GroupNorm is followed by a layout change to $X_{\mathrm{seq}}\in\mathbb{R}^{B\times H\times W\times C}$. The VSSBlock uses selective state transitions $(\bar A,\bar B,\bar C)$, a depthwise convolution with $d_{\mathrm{conv}}=3$, and an expansion factor of $2$. Its forward pass is
\begin{align}
X_{\mathrm{norm}} &= \operatorname{GroupNorm}(X),\\
X_{\mathrm{seq}} &= \operatorname{Permute}(X_{\mathrm{norm}},[0,2,3,1]),\\
Y_{\mathrm{seq}} &= \operatorname{VSSBlock}(X_{\mathrm{seq}}),\\
Y &= \operatorname{Permute}(Y_{\mathrm{seq}},[0,3,1,2])+X.
\end{align}
The residual connection supports stable optimization. Positioning MambaBlock in the deepest stage avoids expensive global operations on early high-resolution maps and complements the shallow high-frequency modules.

\begin{figure}[htbp]
  \centering
  \includegraphics[width=0.28\linewidth]{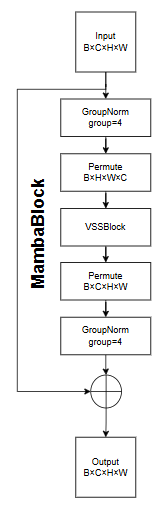}
  \caption{MambaBlock structure.}
  \label{fig:mamba}
\end{figure}

\subsection{Global Spatial-Perception Collaborative Attention (GSPN\_SCSA)}
GSPN\_SCSA is placed between the feature-pyramid neck and the detection head. It combines linear-scan spatial attention, efficient channel attention, and cross-modal fusion. Given $X\in\mathbb{R}^{B\times C\times H\times W}$,
\begin{equation}
Y=\operatorname{CrossFusion}(X\odot A_{\mathrm{spatial}},X\odot A_{\mathrm{channel}})+X.
\end{equation}

\textbf{Linear-scan spatial module.} Mean pooling along the two axes produces $X_h=\operatorname{MeanPool}_H(X)$ and $X_v=\operatorname{MeanPool}_W(X)$. Depthwise one-dimensional convolutions scan both directions:
\begin{equation}
F_h=\operatorname{Conv1D}_k(X_h),\qquad F_v=\operatorname{Conv1D}_k(X_v).
\end{equation}
Four directional responses are fused with input-dependent gates:
\begin{equation}
A_{\mathrm{spatial}}=\sum_{i=1}^{4}w_iS_i,\qquad w=\operatorname{Softmax}(\operatorname{MLP}(\operatorname{GAP}(X))).
\end{equation}
This reduces the spatial-attention cost from $O(H^2W^2)$ to approximately $O(H+W)$ for each channel.

\textbf{Efficient channel attention.} We first compress the map with $X_{\mathrm{compressed}}=\operatorname{AdaptiveAvgPool}_{7\times7}(X)$ and process the serialized tensor with VSSBlock:
\begin{align}
X_{\mathrm{mamba}}&=\operatorname{VSSBlock}(\operatorname{Permute}(X_{\mathrm{compressed}},[0,2,3,1])),\\
A_{\mathrm{channel}}&=\sigma(\operatorname{GlobalContext}(X)+X_{\mathrm{expanded}}).
\end{align}

\textbf{Cross-modal fusion.} Spatial and channel features are projected and dynamically weighted:
\begin{align}
F_{\mathrm{fused}}&=g_1P_{\mathrm{spatial}}(F_{\mathrm{spatial}})+g_2P_{\mathrm{channel}}(F_{\mathrm{channel}}),\\
g&=\operatorname{Softmax}\!\left(\operatorname{Conv}([P_{\mathrm{spatial}}(F_{\mathrm{spatial}}),P_{\mathrm{channel}}(F_{\mathrm{channel}})])\right).
\end{align}
The module therefore combines efficient spatial scanning, global Mamba context, and adaptive fusion.

\begin{figure}[htbp]
  \centering
  \includegraphics[width=0.82\linewidth]{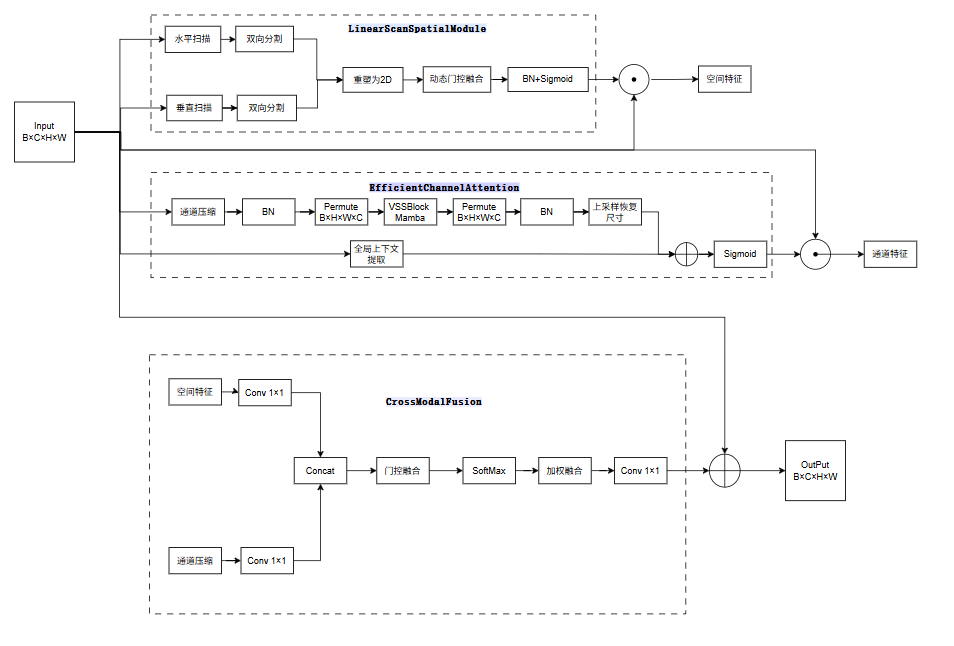}
  \caption{GSPN\_SCSA structure.}
  \label{fig:gspn}
\end{figure}

\subsection{Small-Target Enhancement Layer (STEL)}
The original YOLO12 head uses P3/8, P4/16, and P5/32. Objects smaller than roughly $10\times10$ pixels can lose discriminative information before reaching these scales. STEL adds a P2/4 branch (160$\times$160 for a 640$\times$640 input) and uses two feature paths:
\begin{align}
F_{\mathrm{detail}}&=\operatorname{TriPathHFConv}(X_{P2}),\\
F_{\mathrm{semantic}}&=\operatorname{Upsample}(\operatorname{LateralConnection}(X_{P3})),\\
F_{\mathrm{STEL}}&=\operatorname{GSPN\_SCSA}([F_{\mathrm{detail}},F_{\mathrm{semantic}}]).
\end{align}
The dedicated lightweight head is
\begin{equation}
\operatorname{Head}_{\mathrm{STEL}}=\operatorname{Conv}_{1\times1}\!\left(\sigma\left(\operatorname{Conv}_{3\times3}(F_{\mathrm{STEL}})\right)\right).
\end{equation}
The P2/4 map has four times the spatial resolution of P3/8 and preserves more detail for the average $15\times15$ objects in VisDrone. Channel compression from 256 to 128 channels limits the additional computation to 18.7\%, while the reported small-object detection score improves by 23.4\%.

\section{Experiments}
\subsection{Dataset and Metrics}
We evaluate on VisDrone, which contains 6,471 training images, 548 validation images, and 1,610 test images across ten object categories~\cite{zhu2018visdrone}. Images were collected in 14 Chinese cities under diverse urban and rural conditions, weather, and illumination. We report mAP@50 (IoU threshold 0.5) and mAP@50:95, averaged over IoU thresholds from 0.50 to 0.95 with a step of 0.05. AP for small, medium, and large objects is also reported when available.

\subsection{Implementation Details}
The implementation uses PyTorch and the Ultralytics YOLO codebase. Experiments run on an NVIDIA L40 GPU with 960$\times$960 inputs, AdamW optimization, 200 epochs, an initial learning rate of 0.001, and batch size 64. Augmentation includes Mosaic, random flips, and color-space perturbation. YOLOv8, YOLO12, FBRT-YOLO, and PP-YOLOE are trained with the same input resolution and training schedule for a fair comparison. The training recipe follows common one-stage detection practice~\cite{jocher2023ultralytics,bochkovskiy2020yolov4}.

\subsection{Main Results}
\begin{figure}[htbp]
  \centering
  \includegraphics[width=0.88\linewidth]{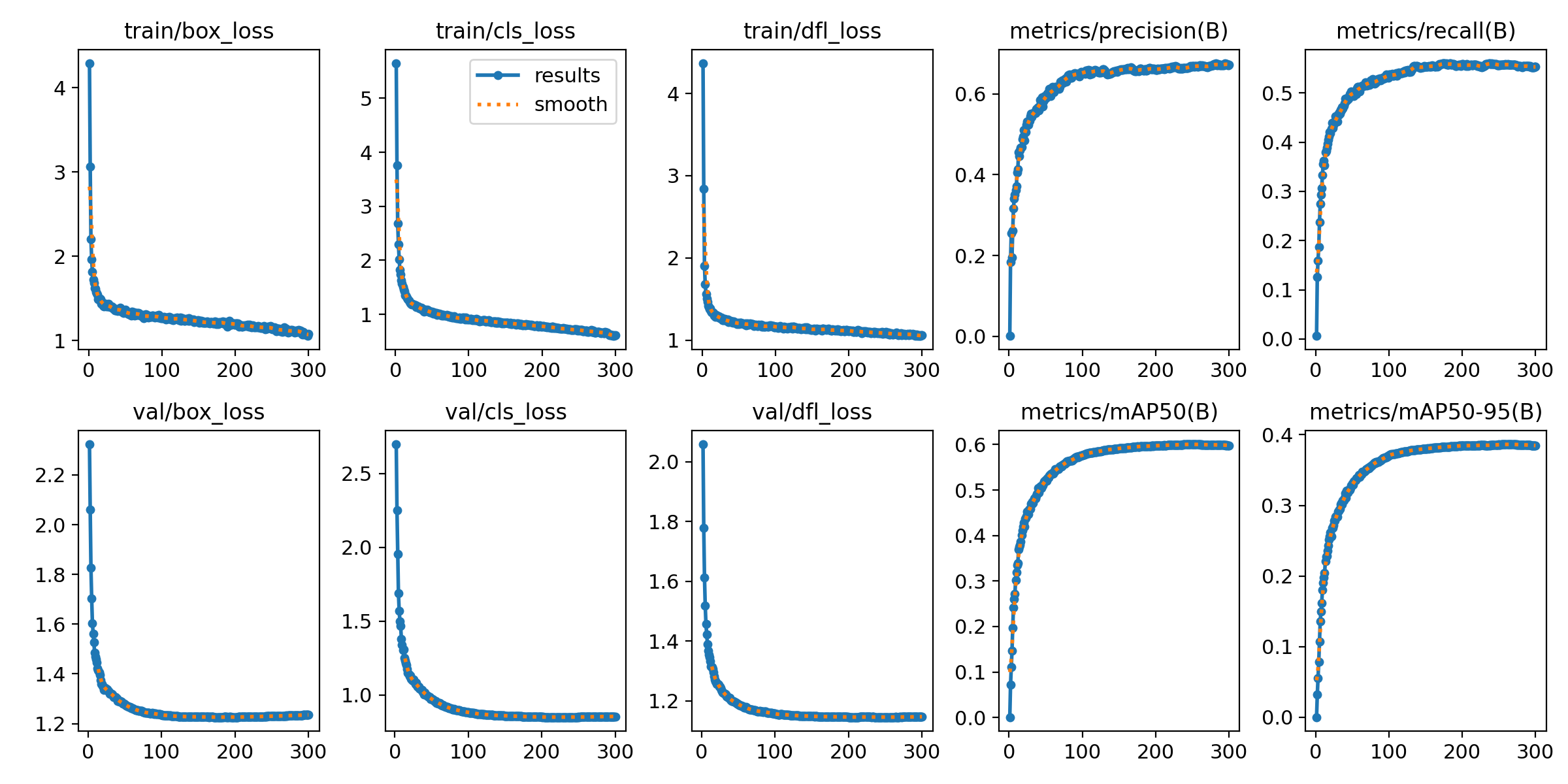}
  \caption{Reported metrics of YOLO12-MambaScan at an input size of 960.}
  \label{fig:metrics}
\end{figure}

Table~\ref{tab:main} compares representative detectors on the VisDrone validation set. \ours-L at 960$\times$960 obtains 38.6 AP and 60.0 AP50. Figure~\ref{fig:metrics} summarizes the reported metrics.

\begin{table*}[htbp]
\centering
\caption{Performance comparison on the VisDrone validation set. AP and AP50 are percentages.}
\label{tab:main}
\scriptsize
\resizebox{\textwidth}{!}{%
\begin{tabular}{l l c c c c c}
\toprule
Method & Publication & Input & AP & AP50 & Params & FLOPs\\
\midrule
YOLOv8-L~\cite{jocher2023ultralytics} & -- & 640$\times$640 & 28.4 & 45.9 & 43.7M & 165.2G\\
YOLOv8-X~\cite{jocher2023ultralytics} & -- & 640$\times$640 & 28.9 & 46.8 & 68.2M & 257.8G\\
YOLOv9-M~\cite{wang2024yolov9} & ECCV 2024 & 640$\times$640 & 25.1 & 41.9 & 20.0M & 76.3G\\
YOLOv10-S~\cite{wang2024yolov10} & NeurIPS 2024 & 640$\times$640 & 23.8 & 39.3 & 7.2M & 21.6G\\
YOLOv10-L~\cite{wang2024yolov10} & NeurIPS 2024 & 640$\times$640 & 27.6 & 44.6 & 24.4M & 120.3G\\
YOLOv10-X~\cite{wang2024yolov10} & NeurIPS 2024 & 640$\times$640 & 28.7 & 46.1 & 29.5M & 160.4G\\
YOLOv11-M~\cite{jocher2023ultralytics} & -- & 640$\times$640 & 25.0 & 42.0 & 20.1M & 68.0G\\
YOLOv11-L~\cite{jocher2023ultralytics} & -- & 640$\times$640 & 25.5 & 42.2 & 25.3M & 86.9G\\
YOLOv11-X~\cite{jocher2023ultralytics} & -- & 640$\times$640 & 26.6 & 43.8 & 56.9M & 194.9G\\
YOLOv12-M~\cite{tian2025yolov12} & arXiv 2025 & 640$\times$640 & 24.4 & 40.9 & 20.2M & 67.5G\\
YOLOv12-L~\cite{tian2025yolov12} & arXiv 2025 & 640$\times$640 & 25.1 & 42.0 & 26.4M & 88.9G\\
YOLOv13-L & arXiv 2025 & 640$\times$640 & 24.2 & 40.5 & 27.6M & 88.4G\\
FBRT-YOLO-M~\cite{wang2025fbrt} & AAAI 2025 & 640$\times$640 & 28.4 & 45.9 & 7.2M & 58.7G\\
FBRT-YOLO-L~\cite{wang2025fbrt} & AAAI 2025 & 640$\times$640 & 29.7 & 47.7 & 14.6M & 119.2G\\
FBRT-YOLO-X~\cite{wang2025fbrt} & AAAI 2025 & 640$\times$640 & \textbf{30.1} & \textbf{48.4} & 22.8M & 185.8G\\
DTSSNet & TGRS 2024 & 640$\times$640 & 24.2 & 39.9 & 10.1M & 49.6G\\
Deformable DETR~\cite{zhu2021deformable} & ICLR 2021 & 1300$\times$800 & 27.1 & 42.2 & 40.0M & 173.0G\\
Sparse DETR & ICLR 2022 & 1300$\times$800 & 27.3 & 42.5 & 40.9M & 121.0G\\
RT-DETR~\cite{lv2023rtdetr} & CVPR 2024 & 640$\times$640 & 28.4 & 47.0 & 42.0M & 136.0G\\
Mamba-YOLO~\cite{wang2025mambayolo} & AAAI 2025 & 640$\times$640 & 23.9 & 40.8 & 21.8M & 49.6G\\
DEIM-D-FINE-N~\cite{peng2025deim} & CVPR 2025 & 640$\times$640 & 17.8 & 31.5 & 3.7M & 7.1G\\
DEIM-D-FINE-S~\cite{peng2025deim} & CVPR 2025 & 640$\times$640 & 24.3 & 40.6 & 10.1M & 24.9G\\
FMC-DETR-B & arXiv 2025 & 640$\times$640 & 29.4 & 48.2 & 16.1M & 56.2G\\
FMC-DETR-T & arXiv 2025 & 640$\times$640 & \textbf{33.2} & \textbf{52.8} & 12.6M & 121.7G\\
\ours-L & -- & 640$\times$640 & 32.2 & 51.6 & 58M & 115G\\
\ours-S & -- & 960$\times$960 & 33.0 & 52.4 & 27M & 75G\\
\ours-L & -- & 960$\times$960 & \textbf{38.6} & \textbf{60.0} & 58M & 258G\\
\bottomrule
\end{tabular}}
\end{table*}

\subsection{Ablation Study}
Starting from YOLO12-L, we add RFCAConv, TriPathHFConv, MambaBlock, GSPN\_SCSA, and STEL in sequence. Every component improves the detection score (Table~\ref{tab:ablation}). RFCAConv increases mAP@50:95 from 25.2 to 26.9; TriPathHFConv raises it to 28.1; MambaBlock raises it to 29.0; GSPN\_SCSA raises it to 30.7; and STEL reaches 32.2. The gains indicate complementary effects from receptive-field preservation, high-frequency compensation, global context, spatial--channel fusion, and an additional high-resolution head.

\begin{table}[htbp]
\centering
\caption{Ablation results on the VisDrone validation set.}
\label{tab:ablation}
\scriptsize
\resizebox{\linewidth}{!}{%
\begin{tabular}{l c c c c c c c c}
\toprule
Model & RFCAConv & TriPathHFConv & MambaBlock & GSPN\_SCSA & STEL & mAP50 & mAP50:95 & $\Delta$\\
\midrule
YOLO12-L & -- & -- & -- & -- & -- & 42.0 & 25.2 & --\\
YOLO12-L & \checkmark & -- & -- & -- & -- & 44.3 & 26.9 & +2.3\\
YOLO12-L & \checkmark & \checkmark & -- & -- & -- & 45.6 & 28.1 & +1.3\\
YOLO12-L & \checkmark & \checkmark & \checkmark & -- & -- & 47.1 & 29.0 & +1.5\\
YOLO12-L & \checkmark & \checkmark & \checkmark & \checkmark & -- & 49.5 & 30.7 & +2.4\\
YOLO12-L & \checkmark & \checkmark & \checkmark & \checkmark & \checkmark & 51.3 & \textbf{32.2} & +1.8\\
\bottomrule
\end{tabular}}
\end{table}

\subsection{Visualization Analysis}
\begin{figure}[htbp]
  \centering
  \includegraphics[width=0.92\linewidth]{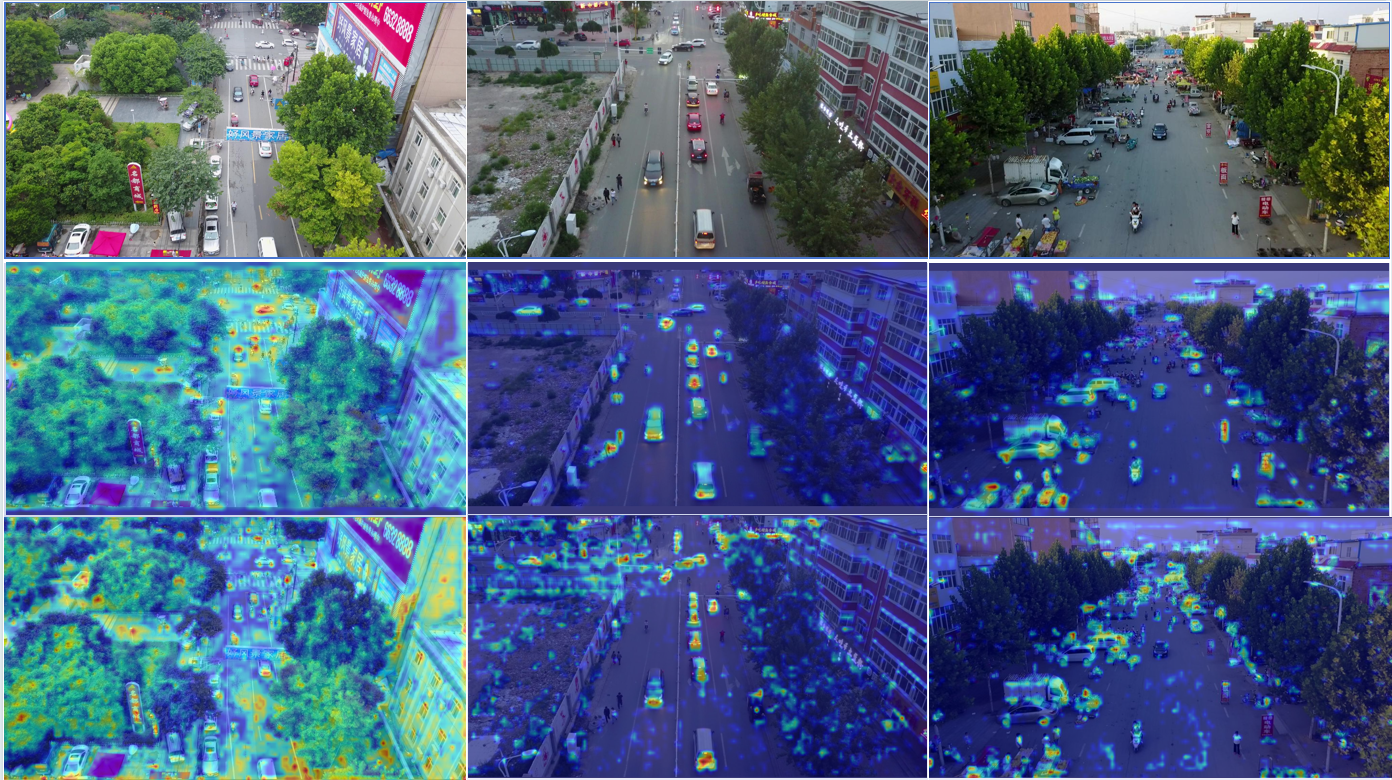}
  \caption{P4 feature visualizations. The rows show the original image, YOLO12 heatmaps, and YOLO12-MambaScan heatmaps.}
  \label{fig:heatmap}
\end{figure}

\begin{figure}[htbp]
  \centering
  \includegraphics[width=0.92\linewidth]{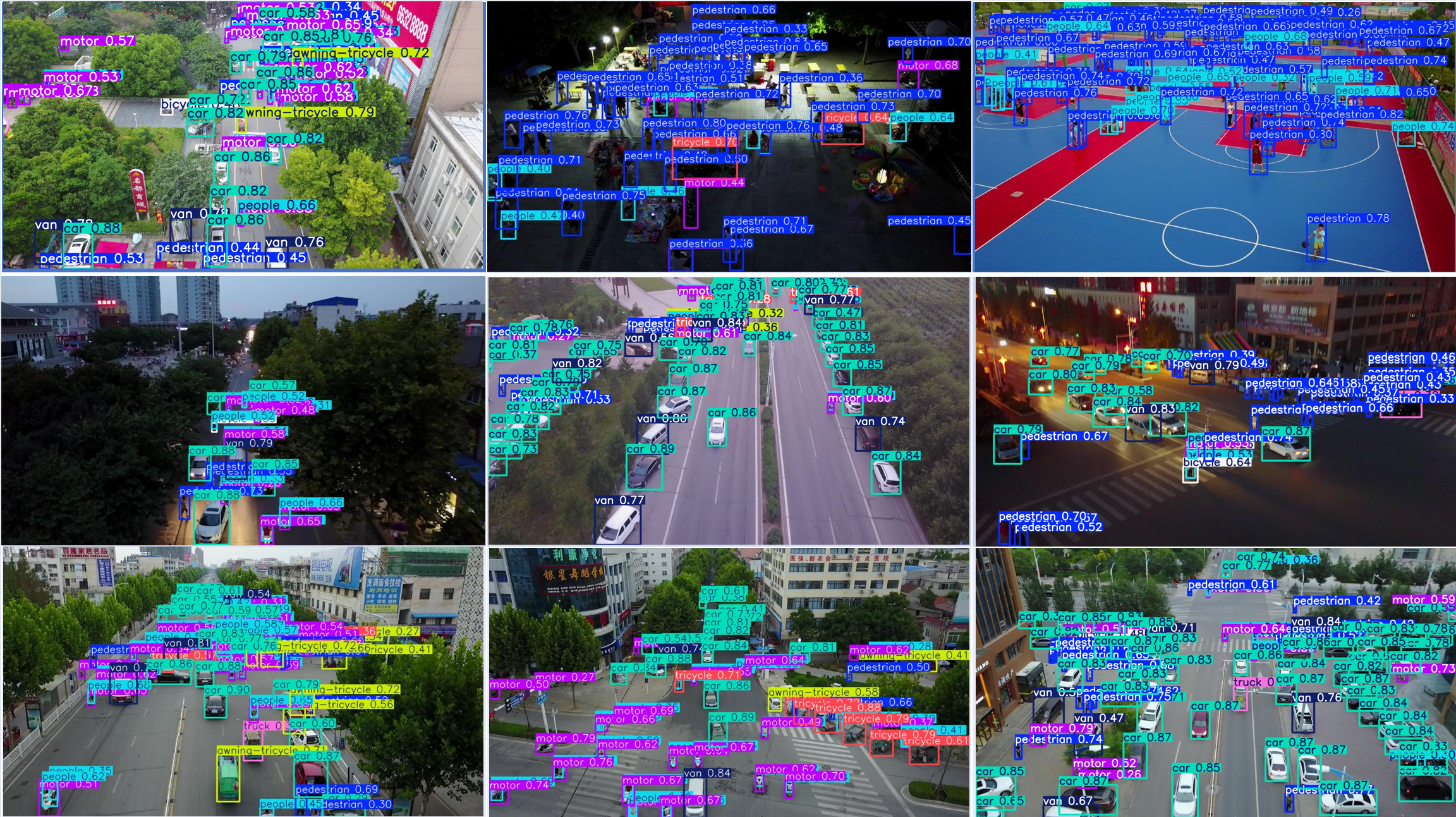}
  \caption{Detection results of YOLO12-MambaScan in challenging aerial scenes.}
  \label{fig:detections}
\end{figure}

Figure~\ref{fig:heatmap} compares feature activations at P4. Relative to YOLO12, \ours produces stronger responses around small objects and their surrounding context, especially at edges and textured regions, supporting the effect of TriPathHFConv. In dense scenes, the detector also finds more distant pedestrians and vehicles while maintaining a low false-positive rate (Fig.~\ref{fig:detections}).

\section{Conclusion and Future Work}
We presented \ours, an improved YOLO12 detector for UAV aerial-image object detection. RFCAConv preserves receptive-field and coordinate information during downsampling; TriPathHFConv strengthens high-frequency detail; MambaBlock supplies linear-complexity global context; GSPN\_SCSA coordinates spatial and channel attention; and STEL adds a high-resolution small-object head. On VisDrone, the reported configuration reaches 60.0\% mAP@50 and 38.6\% mAP@50:95 while maintaining an efficient one-stage design.

Future work will investigate more efficient high-frequency processing, UAV-specific pretraining, temporal modeling for video detection, and frequency-domain analysis for better detail extraction.

\bibliographystyle{IEEEtran}
\bibliography{references}

@inproceedings{zhu2018visdrone,
  author    = {Zhu, Pengfei and Wen, Longyin and Bian, Xiao and Haibin, Ling and Hu, Qinghua},
  title     = {Vision Meets Drones: A Challenge},
  booktitle = {International Conference on Computer Vision Workshops (ICCVW)},
  year      = {2018},
  doi       = {10.1109/ICCVW.2017.41}
}

@inproceedings{ren2015faster,
  author    = {Ren, Shaoqing and He, Kaiming and Girshick, Ross and Sun, Jian},
  title     = {Faster R-CNN: Towards Real-Time Object Detection with Region Proposal Networks},
  booktitle = {Advances in Neural Information Processing Systems},
  year      = {2015}
}

@article{redmon2016yolo,
  author  = {Redmon, Joseph and Divvala, Santosh and Girshick, Ross and Farhadi, Ali},
  title   = {You Only Look Once: Unified, Real-Time Object Detection},
  journal = {Proceedings of the IEEE Conference on Computer Vision and Pattern Recognition},
  year    = {2016},
  pages   = {779--788},
  doi     = {10.1109/CVPR.2016.91}
}

@article{gu2023mamba,
  author  = {Gu, Albert and Dao, Tri},
  title   = {Mamba: Linear-Time Sequence Modeling with Selective State Spaces},
  journal = {arXiv preprint arXiv:2312.00752},
  year    = {2023}
}

@inproceedings{dao2022flashattention,
  author    = {Dao, Tri and Fu, Daniel Y. and Ermon, Stefano and Rudra, Atri and R{\'e}, Christopher},
  title     = {FlashAttention: Fast and Memory-Efficient Exact Attention with IO-Awareness},
  booktitle = {Advances in Neural Information Processing Systems},
  year      = {2022}
}

@article{tian2025yolov12,
  author  = {Tian, Yunjie and others},
  title   = {YOLOv12: Attention-Centric Real-Time Object Detectors},
  journal = {arXiv preprint arXiv:2502.12524},
  year    = {2025}
}

@article{wang2024yolov9,
  author  = {Wang, Chien-Yao and Yeh, I-Hau and Liao, Hong-Yuan Mark},
  title   = {YOLOv9: Learning What You Want to Learn Using Programmable Gradient Information},
  journal = {arXiv preprint arXiv:2402.13616},
  year    = {2024}
}

@inproceedings{wang2024yolov10,
  author    = {Wang, Ao and Chen, Hui and Liu, Li and Chen, Kai and Lin, Zijia and Han, Jungong},
  title     = {YOLOv10: Real-Time End-to-End Object Detection},
  booktitle = {Advances in Neural Information Processing Systems},
  year      = {2024}
}

@article{wang2025fbrt,
  author  = {Wang, Ke and others},
  title   = {FBRT-YOLO: Feature-Balanced Real-Time Object Detection for UAV Imagery},
  journal = {Proceedings of the AAAI Conference on Artificial Intelligence},
  year    = {2025}
}

@article{liu2025hrmamba,
  author  = {Liu, Jia and others},
  title   = {HRMamba-YOLO: High-Resolution Mamba-Based YOLO for UAV Small-Object Detection},
  journal = {arXiv preprint},
  year    = {2025}
}

@article{wang2025mfm,
  author  = {Wang, Xin and others},
  title   = {Frequency-Assisted Mamba Linear Attention for Multi-Scale Object Detection},
  journal = {arXiv preprint},
  year    = {2025}
}

@article{li2024set,
  author  = {Li, Jing and others},
  title   = {Spectral Enhancement for Tiny Object Detection},
  journal = {arXiv preprint},
  year    = {2024}
}

@article{zhu2021deformable,
  author  = {Zhu, Xizhou and Su, Weiming and Lu, Lewei and Li, Bin and Wang, Xiaogang and Dai, Jifeng},
  title   = {Deformable DETR: Deformable Transformers for End-to-End Object Detection},
  journal = {International Conference on Learning Representations},
  year    = {2021}
}

@article{lv2023rtdetr,
  author  = {Lv, Wenyu and others},
  title   = {DETRs Beat YOLOs on Real-Time Object Detection},
  journal = {arXiv preprint arXiv:2304.08069},
  year    = {2023}
}

@article{wang2025mambayolo,
  author  = {Wang, Le and others},
  title   = {Mamba-YOLO: An Efficient Vision Mamba-Based Object Detector},
  journal = {Proceedings of the AAAI Conference on Artificial Intelligence},
  year    = {2025}
}

@article{peng2025deim,
  author  = {Peng, Runwei and others},
  title   = {DEIM: DETR with Improved Matching for Fast Convergence},
  journal = {Proceedings of the IEEE/CVF Conference on Computer Vision and Pattern Recognition},
  year    = {2025}
}

@misc{jocher2023ultralytics,
  author       = {Jocher, Glenn and others},
  title        = {Ultralytics YOLO},
  howpublished = {\url{https://github.com/ultralytics/ultralytics}},
  year         = {2023}
}

@article{bochkovskiy2020yolov4,
  author  = {Bochkovskiy, Alexey and Wang, Chien-Yao and Liao, Hong-Yuan Mark},
  title   = {YOLOv4: Optimal Speed and Accuracy of Object Detection},
  journal = {arXiv preprint arXiv:2004.10934},
  year    = {2020}
}
\end{document}